\documentclass[10pt,twocolumn,letterpaper]{article}

\usepackage{cvpr}              
\usepackage[accsupp]{axessibility}  

\usepackage{tabularx}

\usepackage{graphicx}
\usepackage{amsmath}
\usepackage{amssymb}
\usepackage{booktabs}
\newcolumntype{Y}{>{\centering\arraybackslash}X}
\usepackage{pifont}
\newcommand{\cmark}{\ding{51}}%
\newcommand{\xmark}{\ding{55}}%
\newcommand{\rev}[1]{{\textcolor{black}{ #1}}}

\usepackage{amsmath}
\usepackage{color}
\usepackage{soul}
\usepackage{multirow}
\usepackage{amssymb}
\usepackage{wrapfig}
\usepackage{mathtools}
\usepackage[ruled,linesnumbered]{algorithm2e}
\usepackage{esvect}
\usepackage{ textcomp }

\SetKwRepeat{Do}{do}{while}%

\definecolor{red}{rgb}{1.0, 0.0, 0.0}
\definecolor{green}{rgb}{0.0, 1.0, 0.0}
\definecolor{blue}{rgb}{0.0, 0.0, 1.0}
\definecolor{purple}{rgb}{0.65,0,0.65}
\definecolor{magenta}{rgb}{1.0,0,1.0}
\definecolor{turquoise}{cmyk}{0.65,0,0.1,0.1}
\definecolor{darkgreen}{rgb}{0.0, 0.5, 0.0}
\definecolor{darkred}{rgb}{0.5, 0.0, 0.0}
\definecolor{darkblue}{rgb}{0.0, 0.0, 0.5}
\definecolor{lightsalmon}{rgb}{1.0, 0.62, 0.48}
\definecolor{cyan}{rgb}{0.0, 1.0, 1.0}

\newcommand{\erase}[1]{}

\newcommand{\hide}[1]{{\color{red}-- -- -- --}}

\definecolor{blue}{rgb}{0.0, 0.0, 1.0}

\graphicspath{{figs/}}

\usepackage[pagebackref,breaklinks,colorlinks]{hyperref}

\usepackage[capitalize]{cleveref}
\crefname{section}{Sec.}{Secs.}
\Crefname{section}{Section}{Sections}
\Crefname{table}{Table}{Tables}
\crefname{table}{Tab.}{Tabs.}

\def\cvprPaperID{5700} 
\def\confName{CVPR}
\def\confYear{2023}

\begin{document}

\title{SECAD-Net: Self-Supervised CAD Reconstruction by Learning Sketch-Extrude Operations}

\author{Pu Li$^{1,2}$ \quad Jianwei Guo$^{1,2}$\thanks{Corresponding author: jianwei.guo@nlpr.ia.ac.cn} \quad Xiaopeng Zhang$^{1,2}$ \quad Dong-Ming Yan$^{1,2}$\\
$^1$MAIS, Institute of Automation, Chinese Academy of Sciences\\
$^2$School of Artificial Intelligence, University of Chinese Academy of Sciences
}



\twocolumn[
{%
\renewcommand\twocolumn[1][]{#1}%
\maketitle
\begin{center}
    \centering
    \includegraphics[width=1.0\textwidth]{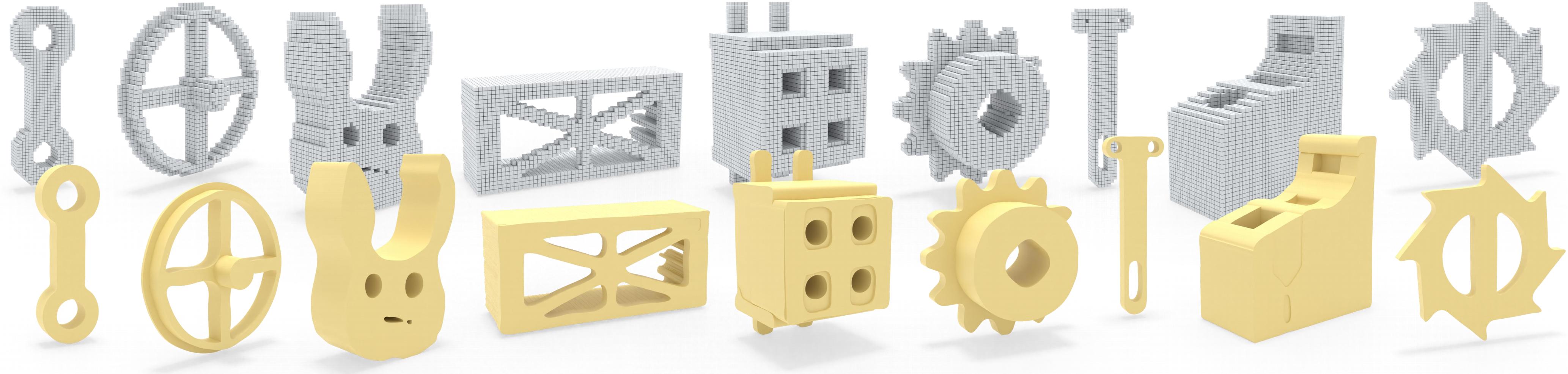}
    \captionof{figure}{\textbf{SECAD-Net for CAD reconstruction}. Starting from a voxel grid (top), SECAD-Net learns \textit{Sketch-Extrude} operations to reconstruct CAD models (bottom), without any supervision of part segmentation and sketch labels. 
    }
\end{center}
}
]

\renewcommand{\thefootnote}{\fnsymbol{footnote}}
\footnotetext[1]{Corresponding author: jianwei.guo@nlpr.ia.ac.cn}

\begin{abstract}
Reverse engineering CAD models from raw geometry is a classic but strenuous research problem. 
Previous learning-based methods rely heavily on labels due to the supervised design patterns or reconstruct CAD shapes that are not easily editable. 
In this work, we introduce \emph{SECAD-Net}, an end-to-end neural network aimed at reconstructing compact and easy-to-edit CAD models in a self-supervised manner. 
Drawing inspiration from the modeling language that is most commonly used in modern CAD software, we propose to learn 2D sketches and 3D extrusion parameters from raw shapes, from which a set of extrusion cylinders can be generated by extruding each sketch from a 2D plane into a 3D body. 
By incorporating the Boolean operation (\ie, union), these cylinders can be combined to closely approximate the target geometry. 
We advocate the use of implicit fields for sketch representation, which allows for creating CAD variations by interpolating latent codes in the sketch latent space.
Extensive experiments on both ABC and Fusion 360 datasets demonstrate the effectiveness of our method, and show superiority over state-of-the-art alternatives including the closely related method for supervised CAD reconstruction. 
We further apply our approach to CAD editing and single-view CAD reconstruction. 
Code will be released at \url{https://github.com/BunnySoCrazy/SECAD-Net}.
\end{abstract}

\section{Introduction}\label{sec:introduction}


CAD reconstruction is one of the most sought-after geometric modeling technologies, which plays a substantial role in reverse engineering in case of the original design document is missing or the CAD model of a real object is not available.  
It empowers users to reproduce CAD models from other representations and supports the designer to create new variations to facilitate various engineering and manufacturing applications. 

The advance in 3D scanning technologies has promoted the paradigm shift from time-consuming and laborious manual dimensions to automatic CAD reconstruction.  
A typical line of works~\cite{requicha1992solid,varady1997reverse,beniere2013comprehensive,buonamici2018reverse} first reconstructs a polygon mesh from the scanned point cloud, then followed by mesh segmentation and primitive extraction to obtain a boundary representation (B-rep). Finally, a CAD shape parser is applied to convert the B-rep into a sequence of modeling operations. 
Recently, inspired by the substantial success of point set learning~\cite{achlioptas2018learning,qi2017pointnet++,wang2019dynamic} and deep 3D representations~\cite{mescheder2019occupancy,park2019deepsdf,chen2019learning}, a number of methods have exploited neural networks to improve the above pipeline, \eg, detecting and fitting primitives to raw point clouds directly~\cite{li2019supervised,le2021cpfn,sharma2020parsenet}. 
A few works (\eg, CSG-Net~\cite{sharma2018csgnet}, UCSG-Net~\cite{kania2020ucsg}, and CSG-Stump~\cite{ren2021csg}) further parse point cloud inputs into a constructive solid geometry (CSG) tree by predicting a set of primitives that are then combined with Boolean operations. 
Although achieving encouraging compact representation, they only output a set of simple primitives with limited types (\eg, planes, cylinders, spheres), which restricts their representation capability for reconstructing complex and more general 3D shapes. CAPRI-Net~\cite{yu2022capri} introduces quadric surface primitives and the difference operation based on BSP-Net~\cite{chen2020bsp} to produce complicated convex and concave shapes via a CSG tree. However, controlling the implicit equation and parameters of quadric primitives is difficult for designers to edit the reconstructed models. Thus, the editability of those methods is quite limited.


In this paper, we develop a novel and versatile deep neural framework, named SECAD-Net, to reconstruct high-quality and editable CAD models. 
Our approach is inspired by the observation that a CAD model is usually designed as a command sequence of operations~\cite{shah1998designing,wu2021deepcad,willis2021fusion,cascaval2022differentiable,yang2022discovering}, \ie, a set of planar 2D sketches are first drawn then extruded into 3D solid shapes for Boolean operations to create the final model. 
At the heart of our approach is to learn the sketch and extrude modeling operations, rather than CSG with parametric primitives. 
To determine the position and axis of each sketch plane, SECAD-Net first learns multiple extrusion boxes to decompose the entire shape into multiple local regions. Afterward, for the local profile in each box, we utilize a fully connected network to learn the implicit representation of the sketch. An extrusion operator is then designed to calculate the implicit expression of the cylinders according to the predicted sketch and extrusion parameters. We finally apply a union operation to assemble all extrusion cylinders into the final CAD model. 



Benefiting from our representation, our approach is flexible and efficient to construct a wide range of 3D shapes.
As the predictions of our method are fully interpretable, it allows users to express their ideas to create variations or improve the design by operating on 2D sketches or 3D cylinders intuitively. 
To summarize, our work makes the following contributions: 
\begin{itemize}
\item We present a novel deep neural network for reverse engineering CAD models with self-supervision, leading to faithful reconstructions that closely approximate the target geometry.
\item SECAD-Net is capable of learning implicit sketches and differentiable extrusions from raw 3D shapes without the guidance of ground truth sketch labels. 
\item Extensive experiments demonstrate the superiority of SECAD-Net through comprehensive comparisons. We also showcase its immediate applications to CAD interpolation, editing, and single-view reconstruction.
\end{itemize}

\section{Related work}
\label{sec:relatedWork}

\noindent\textbf{Neural implicit representation.} 
3D shapes can be represented either \textsl{explicitly} (\eg, point sets, voxels, meshes) or \textsl{implicitly} (\eg, signed-distance functions, indicator functions), each of them comes with its own advantages and drawbacks. 
Recently, there is an explosion of neural implicit representations~\cite{mescheder2019occupancy,park2019deepsdf,chen2019learning} that allow for generating detail-rich 3D shapes by predicting the underlying signed distance fields. 
Thanks to the ability to learn priors over shapes, many deep implicit works have been proposed to solve various 3D tasks, such as shape representation and completion~\cite{sitzmann2019scene,Atzmon2020SAL,chibane2020implicit}, image-based 3D reconstruction~\cite{tulsiani2017multi,xu2019disn,yariv2020multiview}, shape abstraction~\cite{tulsiani2017learning,genova2019learning} and novel view synthesis~\cite{mildenhall2020nerf,dellaert2020neural}. 
Theoretically, any of the above shape representations can be used to represent sketches. However, primitive-based methods usually suppress the ability cap of shape representation. In this work, we choose to fit an implicit sketch  representation using a neural network, and show its superiority over other representations (\eg, BSP~\cite{chen2020bsp}) in the ablation study, see Sec.~\ref{sec:ablations}.




\noindent\textbf{Reverse engineering CAD reconstruction.} 
Over the past decades, reverse engineering has been extensively studied; 
it aims at converting measured data (a surface mesh or a point cloud) into solid 3D models that can be further edited and manufactured by industries. 
Traditional approaches addressing this problem consist of the following tasks: (1) segmentation of the point clouds/meshes~\cite{benkHo2004segmentation,zhang2020blending,Shen2022framework}, (2) fitting of parametric primitives to segmented regions~\cite{schnabel2007efficient,cohen2004variational,yan2012variational}, (3) finishing operations for CAD modeling~\cite{benkHo2001algorithms,langbein2004choosing}. 
Important drawbacks of these conventional methods are the time-consuming process and the requirement of a skilled operator to guide the reconstruction~\cite{buonamici2018reverse}. 

\begin{figure*}[!t]
\centering
  \includegraphics[width=1.0\linewidth]{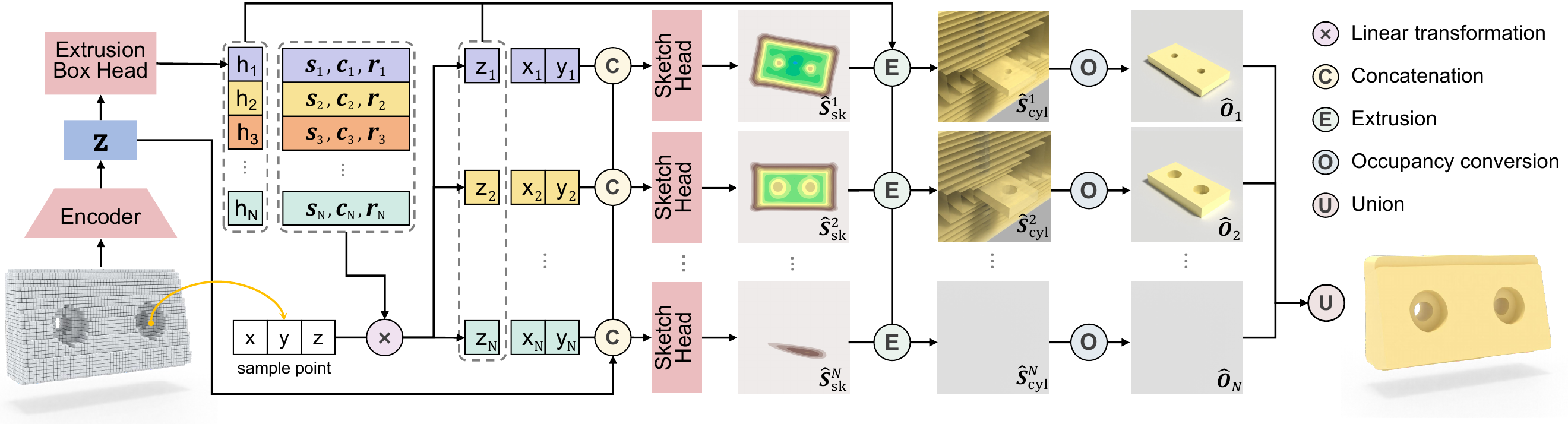}   
  \caption{\textbf{Network architecture for SECAD-Net}: The embedding $\mathbf{z}$ encoded from the voxel input is first fed to the extrusion box head to predict extrusion boxes. It is also sent to the sketch head network to calculate the sketch SDF $\hat{\mathcal{S}}_{\mathsf{sk}}^{i}$ after concatenating with the linear transformed sampling point. $\hat{\mathcal{S}} _{\mathsf{cyl} }^i$ stands for the SDF of the cylinder, which is acquired by extruding $\hat{\mathcal{S}}_{\mathsf{sk}}^{i}$ with height $h_i$. Then we convert $\hat{\mathcal{S}} _{\mathsf{cyl} }^i$ to occupancy of cylinder $\hat{\mathcal{O}}_i$ and finally obtain the complete shape by union all the occupancies. 
  }
  \label{fig:overview}
\end{figure*}

With the release of several large-scale CAD datasets (\eg, ABC~\cite{koch2019abc}, Fusion 360~\cite{willis2021fusion}), SketchGraphs~\cite{seff2020sketchgraphs}), 
numerous approaches have explored deep learning to address primitive segmentation/detection~\cite{yan2021hpnet,le2021cpfn}, parametric curve or surface inference from point clouds~\cite{li2019supervised,sharma2020parsenet,paschalidou2019superquadrics,wang2020pie,guo2022complexgen} or B-rep models~\cite{lambourne2021brepnet,jayaraman2021uv}. However, by only outputting individual curves or surfaces, these methods lack the CAD modeling operations that are needed to build solid models. 
Focusing on CAD generation rather than reconstruction task as ours, some approaches propose deep generative models that predict sequences of CAD modeling operations to produce CAD designs~\cite{li2020sketch2cad,xu2021inferring,wu2021deepcad,willis2021fusion,xu2022skexgen}. 
Aiming at CAD reconstruction involving inverse CSG modeling~\cite{du2018inversecsg}, CSGNet~\cite{sharma2018csgnet} first develops a neural model that parses a shape into a sequence of CSG operations. More recent works follow the line of CSG parsing by advancing the inference without any supervision~\cite{kania2020ucsg}, or improving representation capability with a three-layer reformulation
of the classic CSG-tree~\cite{ren2021csg}, or handling richer geometric and topological variations by introducing quadric surface primitives~\cite{yu2022capri}. While achieving high-quality reconstruction, CSG tends to combine a large number of shape primitives that are not as flexible as the extrusions of 2D sketches and are also not easily user edited to control the final geometry. 

\rev{Motivated by modern design tools, supervised methods are proposed~\cite{uy2022point2cyl,lambourne2022reconstructing} utilizing the sketch-extrude procedural models and learning 2D sketches that can be extruded to 3D shapes. In contrast to their reliance on 2D labels, SECAD-Net is trained in a self-supervised manner.}
\rev{Most closely related to our work is ExtrudeNet~\cite{ren2022extrudenet}. 
SECAD-Net distinguishes itself from ExtrudeNet in several significant aspects:
\romannumeral1) Following the traditional reconstruction process, ExtrudeNet first predicts the parameters of Bézier curves and then converts them into SDFs. In contrast, we jumped out of this paradigm and directly used neural networks to predict the 2D implicit fields of the profiles.
\romannumeral2) ExtrudeNet adopts closed Bézier curves to avoid self-intersection in sketches. This makes ExtrudeNet can only predict star-shaped profiles, which limits the expressive power of their CAD shapes. Our method does not impose any restrictions on the shape of the profile, thus having greater flexibility in shape expression.
\romannumeral3) To pursue the reconstruction effect, ExtrudeNet relies on a larger number of primitives, while our method is able to predict more compact CAD shapes.}

\section{Problem Statement and Overview}

\begin{figure}[!t]
    \centerline{
    \includegraphics[width=1.0\linewidth]{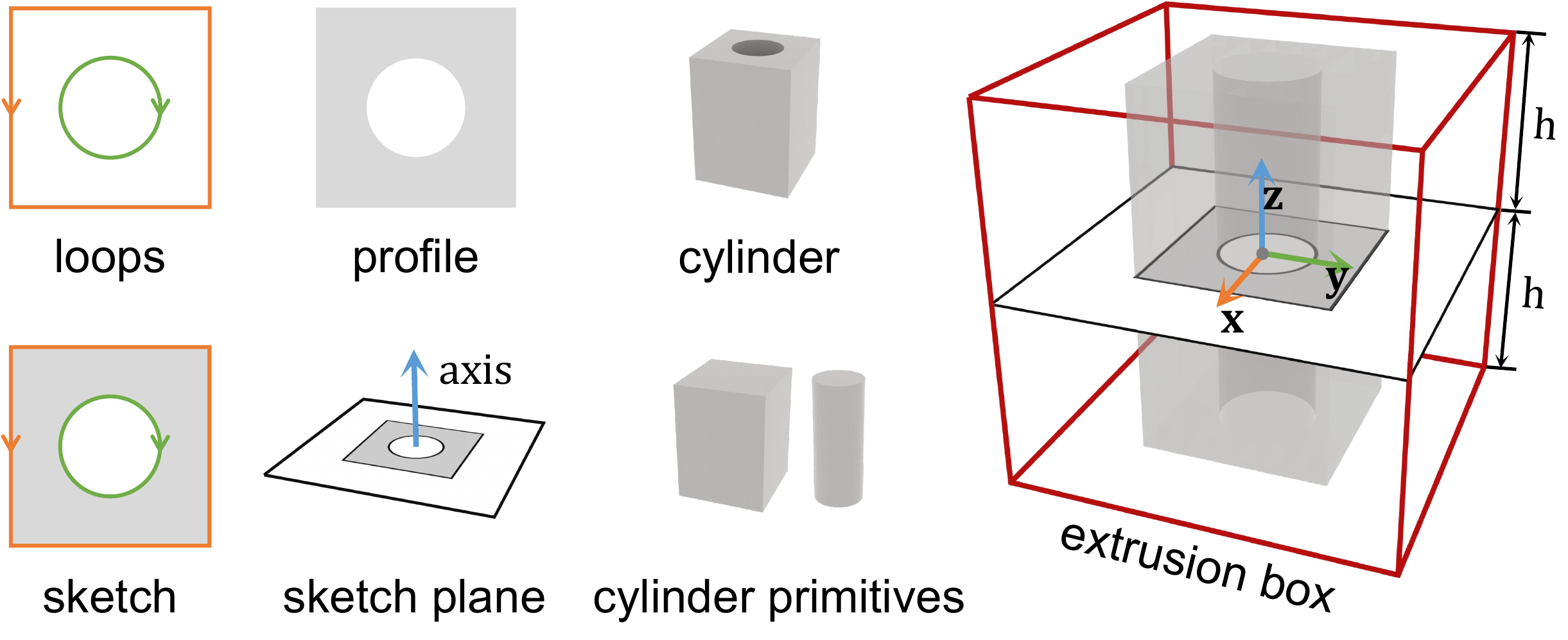}
    }
    \caption{Definitions of CAD terminologies used in this paper. Note that the axis of the sketch plane in the figure is the same as the z-axis in the extrusion box.}
    \label{fig:definition}
\end{figure}

In this section, we present an overview of the proposed approach. To precisely explain our techniques, we first provide the definition of several related terminologies (Fig.~\ref{fig:definition}).

\subsection{Preliminaries}

\noindent\textbf{Definition 1} (Loop, Profile and Sketch) 
\noindent\emph{In CAD terminology, a sketch is represented by a collection of geometric primitives. By referring to a closed curve as a loop and an enclosed region composed of one or multiple \rev{inner/outer} loops as a profile, we define a sketch as the collection of one profile and its loops.}
\\

\noindent\textbf{Definition 2} (Sketch plane and Extrusion box) 
\noindent\emph{A sketch plane is a finite plane with width $w$ and length $l$, containing one or more sketches with the same extrusion height $h$. Then we define an extrusion box as a cuboid with the sketch plane as the base and $2h$ as the height.}
\\

\noindent\textbf{Definition 3} (Cylinder primitive and Cylinder) 
\noindent\emph{In this work, a cylinder refers to the shape obtained by extruding a sketch, and a cylinder primitive is obtained by performing an extrude operation on a closed area formed by a loop. A cylinder may contain one cylinder primitive or be obtained from several cylinder primitives through the Difference operation used in CSG modeling.}

\subsection{Overview}
We formulate the problem of CAD reconstruction as \emph{sketch} and \emph{extrude} inference: taking an input 3D shape, SECAD-Net aims to reconstruct the CAD model by predicting a set of geometric proxies that are decomposed to sketch-extrude operations. 
The overall pipeline of SECAD-Net is visualized in Fig.~\ref{fig:overview}. 
Given a 3D voxel model, we first map it into a latent feature embedding $\mathbf{z}$ by using an encoder based on a 3D convolutional network. 
An extrusion box head network is then applied to predict the parameters of the sketch planes from $\mathbf{z}$. We employ $N$ sketch head network to independently learn $N$ 2D signed distance fields (SDFs) as the implicit representation of a sketch. Next, we design a differentiable extrusion operator to calculate the SDF of the 3D cylinder primitives corresponding to the sketches. Finally, an occupancy transformation operation and a union operation transform the multiple SDFs into the full 3D reconstructed shape as the output of the network. 



\section{Method}
\label{sec:method}
\subsection{Sketch-Extrude Inferring}

We apply a standard 3D CNN encoder to extract a shape code $\mathbf{z}$ with size 256 from the input voxel. The code is then passed to the proposed SECAD-Net to output the sketch and extrusion cylinder parameters. 
Below we introduce the main modules of SECAD-Net following the order of data transmission during prediction.

\noindent\textbf{Extrusion box prediction.} 
We first apply a fully connected layer to predict the parameters of the extrusion boxes. 
Taking the feature encoding $\mathbf{z}$ as input, a decoder, refereed to \emph{sketch box head}, outputs a set of sketch boxes $\mathcal{B}=\left\{\mathbf{s}_i, \mathbf{c}_i,\mathbf{r}_i \mid i \in N  \right\}$, where $\mathbf{s}_i \in \mathbb{R} ^3  $ describes the 2D size (\ie, length and width) of the box, $\mathbf{c}_i \in \mathbb{R} ^3 $ represents the predicted position of the box’s center, and $\mathbf{r}_i \in \mathbb{R}^4$ is the rotation quaternion. The positive z-axis of the extrusion box determines the axial direction $\mathbf{e}_i$ of the sketch plane, and the height of the extrusion box is twice the height of the extrude operation (see Fig.~\ref{fig:definition}).

\noindent\textbf{2D sketch inference.} 
The sketches in each sketch plane depict the shape contained within the sketch box. Inspired by the recent neural implicit shapes~\cite{Atzmon2020SAL,uy2022point2cyl}, we encode the shape of each sketch into a sketch latent space. 
To this end, we first project the 3D sampling points \emph{w.r.t} the corresponding occupancy value onto the sketch plane along the axis $\mathbf{e}_i$.
A \emph{sketch head network} (SK-head) then computes the signed distance from each sampling point to the sketch contour. The distance is negative for points in the sketch and positive for points outside. 
Each SK-head contains $N_{lay}$ layers of fully connected layers, with softplus activation functions used between layers, and we clamp the output distance to [-1,1] in the last layer.
Each 2D point is concatenated with the feature encoding $\mathbf{z}$ as a global condition before being fed into the SK-head. 
Regarding the $i$-th SK-head as an implicit function $f_{i}$, then formally we get: 
\begin{equation}
\hat{\mathcal{S}}_{\mathsf{sk}}^{i} = f_{i}(\mathbf{x}_i^t,\mathbf{z} ),
\end{equation}
where $\mathbf{x}_i^t$ is the result of a linear transformation of the sampling points contained in the $i$-th extrusion box, which can be expressed as $\mathbf{r} _i^{-1}(\mathbf{x}_i -\mathbf{c} _i)$. $\hat{\mathcal{S}}_{\mathsf{sk}}^{i}$ represents the signed distance field of the $i$-th sketch plane.

\noindent\textbf{Differentiable extrusion.}
Next, we calculate the SDF of a cylinder based on the 2D distance field and the extrusion height $h$.
We denote $\Omega$ as the volume between two hyperplanes $p^{u}$ and $p^{l}$, where $p^{u}$ and $p^{l}$ are the upper and lower surfaces on which the cylinder is located. Similarly, we define $\Psi$ as the volume inside the infinite cylinder where the side of the cylinder is located. 
The implicit field of the $i$-th cylinder, $\hat{\mathcal{S}} _{\mathsf{cyl} }^i$, is equal to one of the following cases: 
(1) the distance from a point $\mathbf{x}_{i}$ to $p^{u}$ or $p^{l}$, when $\mathbf{x}_{i} \in \Omega \cap  \Psi^\complement$, where superscript $\complement$ stands for complement; 
(2) $\hat{\mathcal{S}}_{\mathsf{sk} }^{i}$, when $\mathbf{x}_{i} \in \Omega^\complement \cap \Psi $; 
(3) the distance from $\mathbf{x}_{i}$ to the intersection curves of the cylinder and hyperplanes, when $\mathbf{x}_{i} \in \Omega^\complement \cap \Psi^\complement$; 
(4) the maximum distance between $\hat{\mathcal{S}}_{\mathsf{sk} }^{i}$ and the point to $p^{u}$ or $p^{l}$, when $\mathbf{x}_{i} \in \Omega^\complement \cap \Psi$. The sub-formulas for each case are as follows:
\begin{equation}
\hat{\mathcal{S}} _{\mathsf{cyl} }^i = \begin{cases}
max( \hat{\mathcal{S}}_{\mathsf{sk} }^{i} , |\mathbf{x}_{i_z}|-h_{i})  
&, (\hat{\mathcal{S}}_{\mathsf{sk} }^{i}\le 0) \wedge  (|\mathbf{x}_{i_z}|\le h_{i}) \\
|\mathbf{x}_{i_z}|-h _{i}  
&, (\hat{\mathcal{S}}_{\mathsf{sk} }^{i}\le 0) \wedge (|\mathbf{x}_{i_z}|>h_{i})\\
\hat{\mathcal{S}}_{\mathsf{sk} }^{i}
&, (\hat{\mathcal{S}}_{\mathsf{sk} }^{i}>0) \wedge (|\mathbf{x}_{i_z}|\le h_{i}) \\
\left \| \hat{\mathcal{S}}_{\mathsf{sk} }^{i} ,(|\mathbf{x}_{i_z}|-h_{i}) \right \| _2  
&, ( \hat{\mathcal{S}}_{\mathsf{sk} }^{i}>0) \wedge (|\mathbf{x}_{i_z}|>h_{i}) \\
\end{cases}
\end{equation}
Combining the above four sub-formulas with the $max$ and $min$ operations, the following result is obtained:
\begin{eqnarray}
\hat{\mathcal{S}} _{\mathsf{cyl} }^i & = &min(max(\hat{\mathcal{S}}_{\mathsf{sk} }^{i},|\mathbf{x}_{i_z}|-h_{i}),0)   \nonumber\\
  & +   & \left \| max({\hat{\mathcal{S}}_{\mathsf{sk} }^{i}},0),max(|\mathbf{x}_{i_z}|-h_{i},0) \right \| _2
\end{eqnarray}

\noindent\textbf{Occupancy conversion and assembly.} 
The occupancy function represents points inside the shape as 1 and points outside the shape as 0, which can be transformed by SDF. Following~\cite{deng2020cvxnet,ren2021csg}, we use the Sigmoid function to perform differentiable transformation operations: 
\begin{equation}
\label{eq:sigmoid}
\hat{\mathcal{O}}_i = Sigmoid(- \eta \cdot \hat{\mathcal{S}} _{\mathsf{cyl} }^i) \text{.}
\end{equation} 
We finally assemble the occupancy $\hat{\mathcal{O}}_i$ of each cylinder to obtain the reconstructed shape. In order to express complex shapes, many works use intersection, union, and difference operations in CSG in the assembly stage~\cite{chen2020bsp,kania2020ucsg,ren2021csg,yu2022capri}. In contrast to them, we only use the union operation, because the extrusion cylinders can naturally represent concave shapes. This helps us avoid designing intricate loss functions or employing multi-stage training strategies without losing the flexibility of reconstructing shape representations. We adopt the Softmax to compute the union operation as it is shown to be effective in avoiding vanishing gradients~\cite{ren2021csg}: 
\begin{equation}
\label{eq:softmax}
\hat{\mathcal{O}}_{total} =\sum_{i}^{N}  Softmax(\varphi \cdot \hat{\mathcal{O}}_{i})\cdot \hat{\mathcal{O}}_{i},
\end{equation} 
where $\varphi$ is the modulating coefficient and $\hat{\mathcal{O}}_{total}$ is the occupancy representation of the final reconstructed shape.


\subsection{Loss Function}
We train SECAD-Net in a self-supervised fashion through the minimization of the sum of two objective terms. 
The supervision signal is mainly quantified by the reconstruction loss, which measures the mean squared error between the predicted shape occupancy $\hat{\mathcal{O}}_{total}$ and the ground truth $\mathcal{O}_{total}^*$:
\begin{equation}
\mathcal{L}_{\textit{recon}}=\mathbb{E}_{x\in \mathbf{X} } \left [ (\hat{\mathcal{O}}_{total} - \mathcal{O}_{total}^*)^2 \right ],
\end{equation} 
where $x$ is a randomly sampled point in the shape volume. 

However, we find that applying only $\mathcal{L}_{\textit{recon}}$ makes the network always learn fragmented cylinders. 
To tackle this problem, we design a 2D sketch loss to facilitate the network to learn the axis of the sketch plane and the complete profile.  
Specifically, each sketch plane cuts the voxel model to form an occupancy cross-section ${\mathcal{O}_{cs}^{i^*}}$. 
We project the 3D sampling points inside the $i$-th extrusion box $\mathcal{B}^i$ onto the sketch plane along the axial direction, and calculate the difference between the occupancy value of the projected points $\hat{\mathcal{O}}_{proj}$ and ground truth ${\mathcal{O}_{cs}^{i^*}}$ :
\begin{equation}
\mathcal{L}_{\textit{sketch}}=\sum_{i=1}^{N}\mathbb{E}_{x\in \mathcal{B} ^i}\left [ (\hat{\mathcal{O}}_{proj}^i - {\mathcal{O}_{cs}^{i^*}})^2  \right ].
\end{equation} 
The overall objective of SECAD-Net is defined as the combination of the above two terms: 
\begin{equation}
\label{eq:loss}
\mathcal{L}_{total}=\mathcal{L}_{\textit{recon}}+ \lambda \mathcal{L}_{\textit{sketch}},
\end{equation} 
where $\lambda$ is a balance factor.

\subsection{CAD Reconstruction}
\label{sec:reconstruction}
The output of SECAD-Net during the training phase is an implicit occupancy function of the 3D shape. In the prediction stage, we reconstruct CAD models by using sketch-extrude operations instead of the marching cubes method.

\begin{figure}[!t]
    \centerline{
    \includegraphics[width=1.0\linewidth]{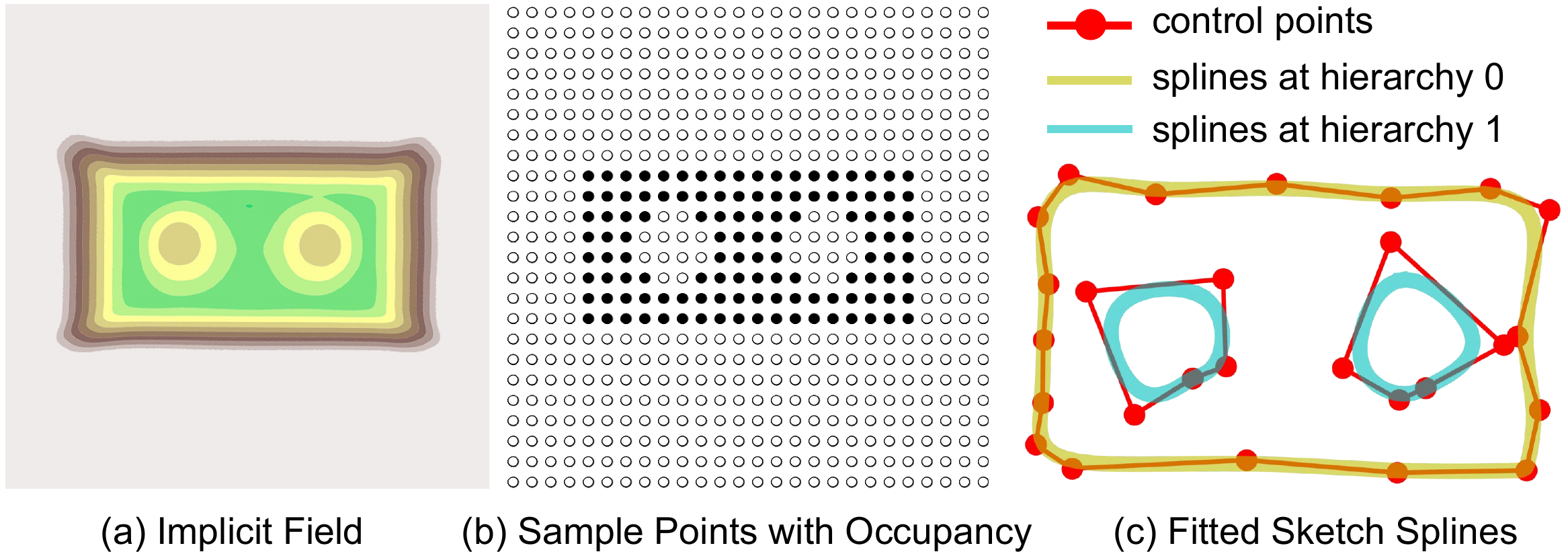}
    }
    \caption{Illustration of converting 2D implicit sketches into the closed B-splines.}
    \label{fig:splines}
\end{figure}

\noindent\textbf{Sketch and extrusion.} 
To convert a 2D implicit field (Fig.~\ref{fig:splines} (a)) in the sketch latent space into an editable sketch, we input uniform 2D sampling points to the SK-head, and attach the implicit value to the position of the sampling point to obtain an explicit image-like 2D profile (Fig.~\ref{fig:splines} (b)). 
We then use the Teh-Chin chain approximation~\cite{teh1989detection} to extract the contours of the profiles and the hierarchical relationships between them. We further apply Dierckx's fitting~\cite{dierckx1982algorithms} to convert the contours into closed B-splines (Fig.~\ref{fig:splines} (c)).

After extruding each sketch to get the cylinder primitives according to half the height of $\mathcal{B}^i$, we assemble cylinder primitives into cylinders by alternately performing union or difference operations according to the hierarchical relationship between contours (primitive at hierarchy 0 \emph{difference} primitives at hierarchy 1 in the case of Fig.~\ref {fig:splines}). Finally, we take the union of all cylinders to obtain the CAD model.

\noindent\textbf{Post-processing.}
We take two post-processing operations to clean up overlapping and shredded shapes in the result. First, for any two cylinders, when their overlapping  coefficient is greater than 0.95, the smaller of them is discarded. Second, we delete all cylinders whose height is less than 0.01 in the reconstruction result. We demonstrate our final reconstructions in Fig.~\ref{fig:compare_ABC} and Fig.~\ref{fig:compare_fusion}.

\subsection{Implementation Details}
SECAD-Net is implemented in PyTorch and trained on a TITAN RTX GPU from NVIDIA\textsuperscript{\textregistered}. 
We train our model using an Adam optimizer~\cite{kingma2015adam} with learning rate $1 \times 10^{-4}$ and beta parameters (0.5, 0.99).
We set both the number of MLP layers in the sketch head network and the number of output cylinders to 4.
For hyper-parameters in Eq.~\ref{eq:sigmoid}, Eq.~\ref{eq:softmax} and Eq.~\ref{eq:loss}, we set $\eta= 150$, $\varphi= 25$ and $\lambda = 0.01$ in default, which generally works well in our experiments.
Employing a similar training strategy to ~\cite{yu2022capri}, we first pre-train SECAD-Net on the training datasets for 1,000 epochs using batch size 24, which takes about 8 hours, and fine-tuning on each test shape for 300 epochs, which takes about 3 minutes per shape.
\section{Experimental Results}
\label{sec:results}

In this section, we examine the performance of SECAD-Net on the ABC dataset~\cite{koch2019abc} and Fusion 360 Gallery~\cite{willis2021fusion}.  
Through extensive comparisons and ablation studies, we demonstrate the effectiveness of our approach and show its superiority over state-of-the-art reference approaches for CAD reconstruction. 

\subsection{Setup}

\noindent\textbf{Dataset preparation.} 
For the ABC dataset, the voxel grids and sampling point data are provided by~\cite{yu2022capri}. We use 5,000 groups of data for training and 1,000 for testing. For Fusion 360, which does not contain available voxels, we first randomly select 6,000 meshes, then discretize them into internally filled voxels. The train-test split is the same as ABC. 
We obtain sampling points with the corresponding occupancy value following~\cite{chen2019learning}. 
The resolution of voxel shapes is $64^3$ for both datasets, and the number of sampling points is 8,192.
Considering that fine-tuning each method and generating high-accuracy meshes is time-consuming, we take 50 shapes from each dataset to form 100 shapes for quantitative evaluation.

\begin{table}[]
\caption{Quantitative comparison between reconstruction results on ABC dataset. }
\label{table:ABC} 
\begin{tabularx}{\columnwidth}{c|Y|Y|Y|Y}
\hline
Methods    & CD↓ & ECD↓ & NC↑ & \#P↓  \\ \hline \hline
UCSG-Net~\cite{kania2020ucsg}   & 1.849 & 1.255 & 0.820 & 12.84 \\ \hline
CSG-Stump~\cite{ren2021csg} & 4.031 & 0.754 & 0.828 & 17.18 \\ \hline
ExtrudeNet~\cite{ren2022extrudenet} & 0.471 & 0.914 & 0.852 & 14.46 \\ \hline
Ours       & \textbf{0.330} & \textbf{0.724} & \textbf{0.863} & \textbf{4.30}   \\ \hline
\end{tabularx}
\end{table}

\begin{table}[]
\caption{Quantitative comparison between reconstruction results on Fusion 360 dataset. }
\label{table:Fusion} 
\begin{tabularx}{\columnwidth}{c|Y|Y|Y|Y}
\hline
Methods    & CD↓ & ECD↓ & NC↑ & \#P↓ \\ \hline \hline
UCSG-Net~\cite{kania2020ucsg}   & 2.950          & 5.277          & 0.770          & 10.84         \\ \hline
CSG-Stump~\cite{ren2021csg} & 2.781          & 4.590          & 0.744          & 12.08         \\ \hline
Point2Cyl~\cite{uy2022point2cyl}  & 13.889         & 14.657         & 0.669          & \textbf{2.76} \\ \hline
ExtrudeNet~\cite{ren2022extrudenet}  & 2.263         & 3.558         & \textbf{0.819}          & 15.72 \\ \hline
Ours       & \textbf{2.052} & \textbf{3.282} & 0.803 & 5.44          \\ \hline
\end{tabularx}
\end{table}

\noindent\textbf{Evaluation metrics.}
For quantitative evaluations, 
we follow the metrics that are commonly used in previous methods~\cite{ren2021csg,yu2022capri}, including symmetric Chamfer Distance ($\textit{CD}$), Normal Consistency ($\textit{NC}$), Edge Chamfer Distance ($\textit{ECD}$). 
Details of computing these metrics are given in the supplemental materials. 
Additionally, we also report the number of generated primitives, $\textit{\#p}$, as a measure of how easy the output CAD results are to edit.

\begin{figure}[!t]
    \centerline{
    \includegraphics[width=1.0\linewidth]{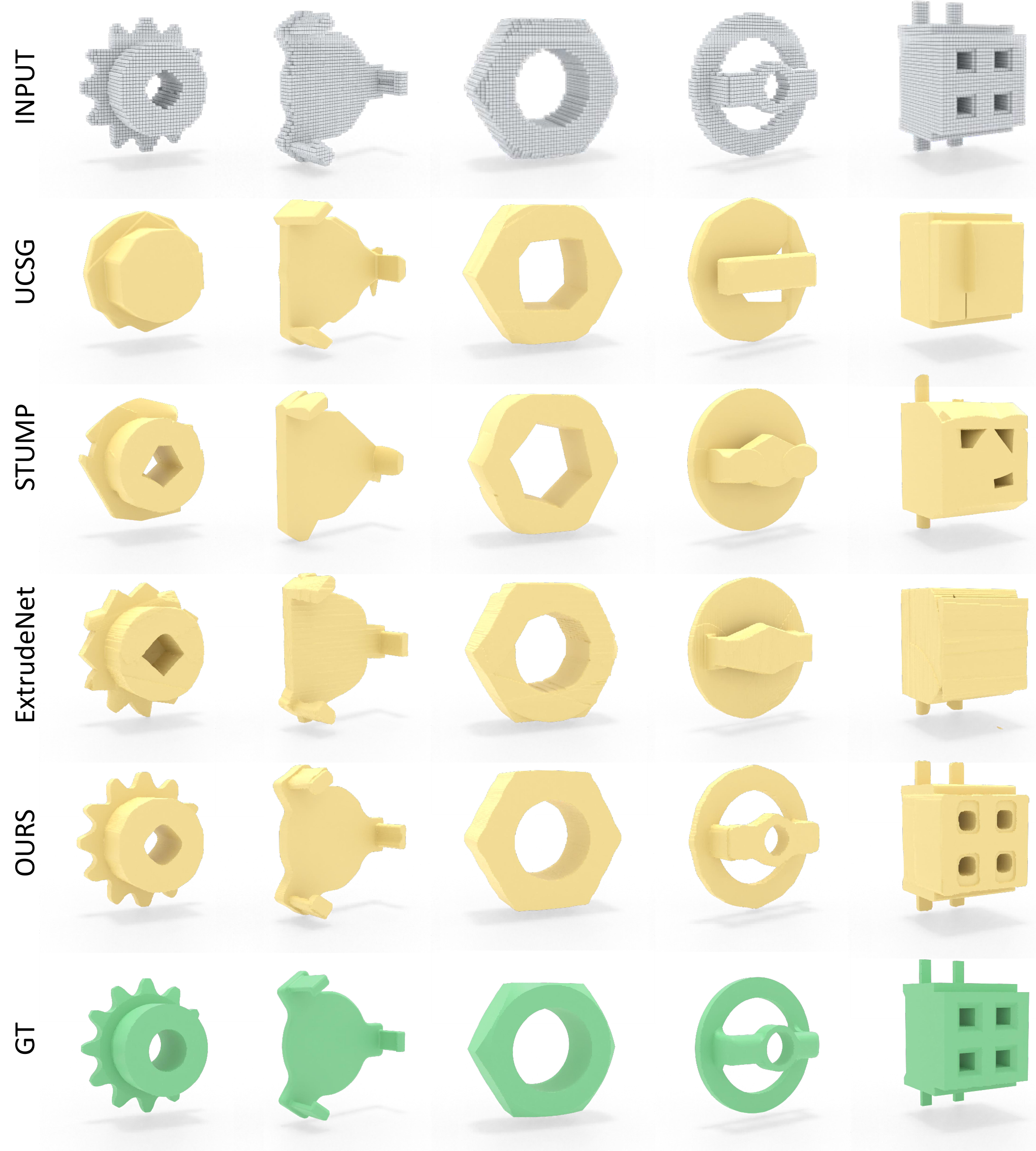}
    }
    \caption{Visual comparison between reconstruction results on ABC dataset.
    }
    \label{fig:compare_ABC}
\end{figure}
\begin{figure}[!t]
    \centerline{
    \includegraphics[width=1.0\linewidth]{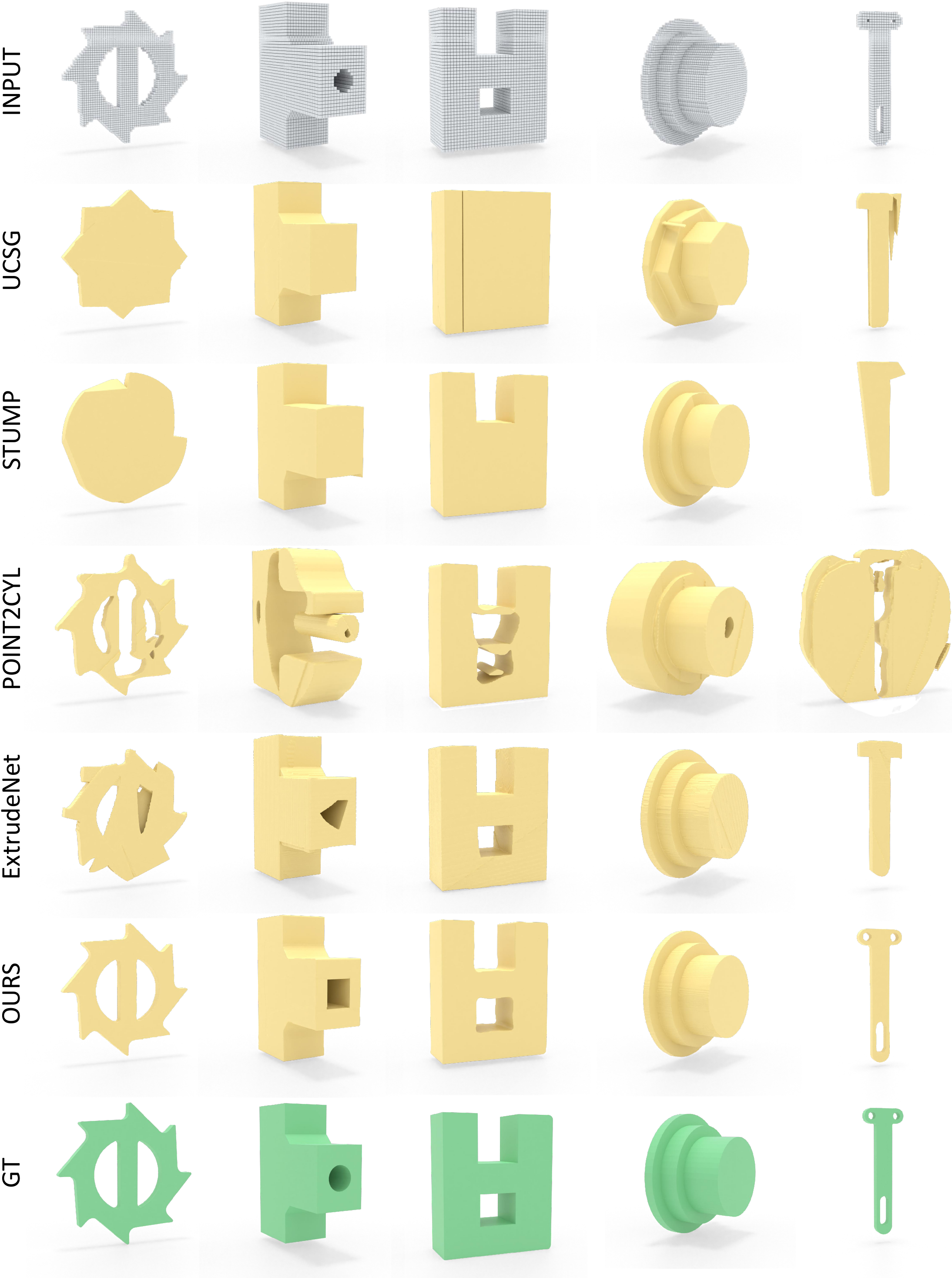}
    }
    \caption{Visual comparison between reconstruction results on Fusion 360 dataset.
    }
    \label{fig:compare_fusion}
\end{figure}

\subsection{Comparison on CAD Reconstruction}

We thoroughly compare our method with two types of primitive-based CAD reconstruction methods that output editable CAD models, including two CSG-like methods (\ie, UCSG-Net~\cite{kania2020ucsg}, CSG-Stump~\cite{ren2021csg}) and \rev{two cylinder decomposition counterpart (\ie, point2Cyl~\cite{uy2022point2cyl}), ExtrudeNet~\cite{ren2022extrudenet}.} 
For each method, we adopt the implementation provided by the corresponding authors, and use the same training strategy for training and fine-tuning.
For CSG-Stump, we set the number of intersection nodes to 64, making it output a comparable number of primitives to other methods.
Those methods provide a plethora of comparisons to other techniques and establish themselves as state-of-the-art. 
Note that for point2Cyl, we only report its results on  Fusion 360 dataset, as the ABC dataset does not provide the labels needed to train point2Cyl. 

Quantitative results on the ABC and Fusion 360 datasets are reported in Table~\ref{table:ABC} and Table~\ref{table:Fusion}, respectively. 
It can be seen that the proposed SECAD-Net outperforms both kinds of methods on all evaluation metrics while still generating a relatively small number of primitives. 
Fig.~\ref{fig:compare_ABC} and Fig.~\ref{fig:compare_fusion} display several qualitative comparison results. For fairness, all the reconstructed CAD models are visualized using marching cubes (MC) with 256 resolution.
Visually as shown in the figures, our method achieves much better geometry and topological fidelity with more accurate structures (\eg, holes, junctions) and sharper features. 


\begin{figure}[!t]
    \centerline{
    \includegraphics[width=1.0\linewidth]{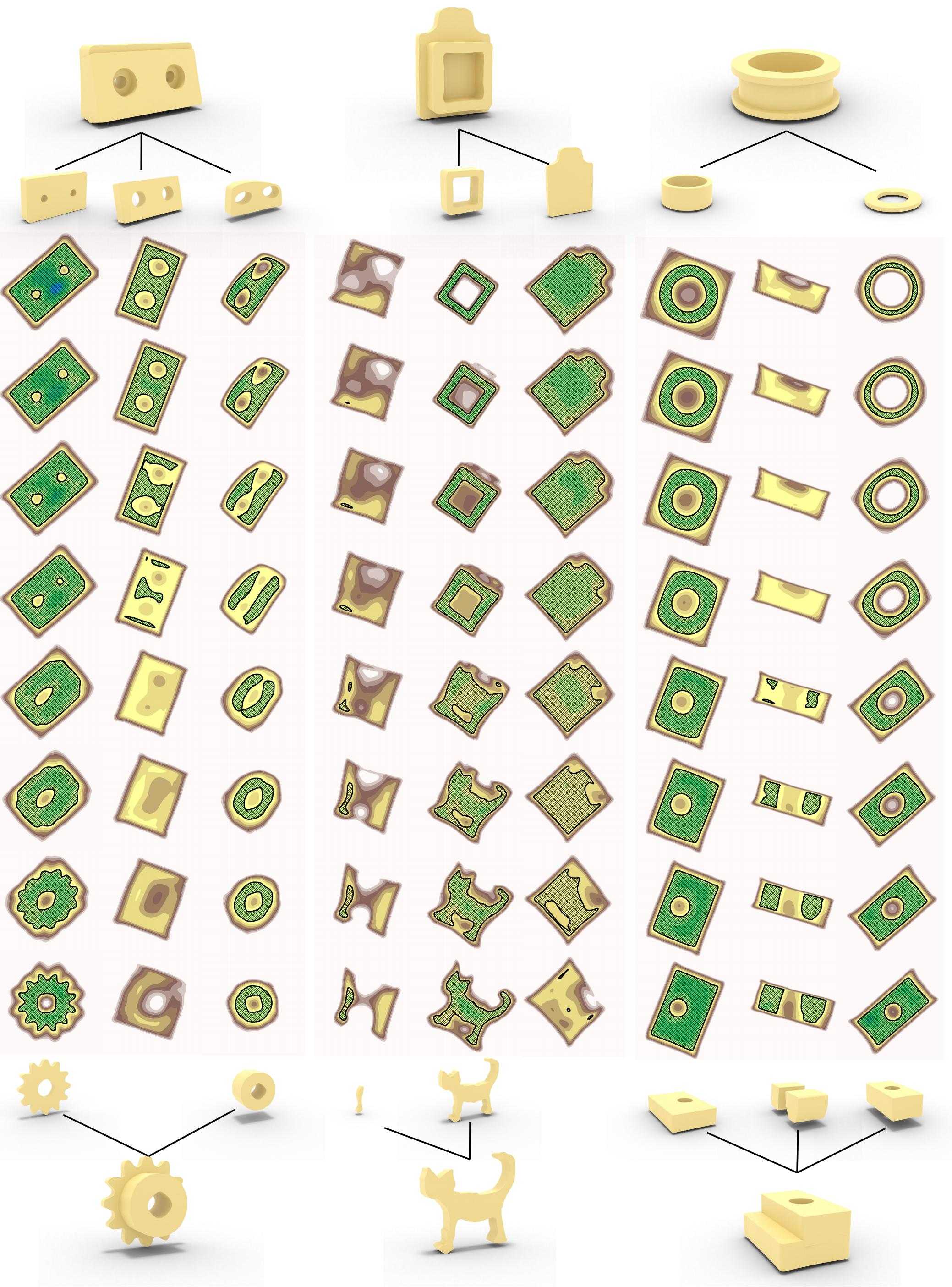}
    }
    \caption{For each example, we encode the sketches of top and bottom shapes in latent vector space and then linearly interpolate the corresponding latent codes.
    }
    \label{fig:sketch_interp}
\end{figure}
\subsection{CAD Generation via Sketch Interpolation}
Although without ground truth labels as guidance, SECAD-Net can learn plausible 2D sketches from raw 3D shapes. 
Thanks to the implicit sketch representation, we are able to generate different CAD variations when a pair of shapes is interpolated in the complete and continuous sketch latent space, as shown in Fig.~\ref{fig:sketch_interp}. The results suggest that the generated sketch is gradually transformed even if the pair of shapes have significantly different structures, and we draw two further conclusions: (1) the predicted position of each extrusion box is relatively deterministic, although the input shape is different (see the left and right column sketches in the leftmost group); (2) when an extrusion box does not contain a shape, our SK-head does not generate a sketch, making the network output an adaptive number of cylinders (see the middle column sketches of the leftmost and the rightmost groups).


\begin{table}[]
\caption{Ablation study on network design and sketch loss. We adopted setting (e) in the final model.}
\label{table:ablation} 
\begin{tabularx}{\columnwidth}{Y|c|c|c|c|c}
\hline
Settings              & (a) & (b) & (c) & (d) & (e) \\ \hline\hline
$N_{sh}$   & 1     & 4     & 4     & 4     & 4     \\ \hline
$N_{lay}$ & 2     & 2     & 2     & 4     & 4     \\ \hline
$N_{cyl}$  & 4     & 4     & 8     & 4     & 4     \\ \hline
$\mathcal{L}_{sketch}$         & \cmark       &  \cmark      &  \cmark      &  \xmark      &  \cmark      \\ \hline 
CD↓                  & 2.627 & 0.993 & 1.504 & 0.336 & \textbf{0.330} \\ \hline
ECD↓                 & 1.754 & 0.882 & 1.098 & 0.772 & \textbf{0.724} \\ \hline
NC↑                  & 0.713 & 0.835 & 0.761 & 0.863 & \textbf{0.863} \\ \hline
\end{tabularx}
\end{table}

\subsection{Ablations}
\label{sec:ablations}
We perform ablation studies to carefully analyze the efficiency of major components of our designed model. All quantitative metrics are measured on the ABC dataset.



\noindent\textbf{Effect of network design and sketch loss.}
We first examine the effect of the number of components/parameters in SECAD-Net, including the number of SK-heads ($N_{sh}$), the number of fully connected layers ($N_{lay}$) in each SK-head, and the number of output cylinders ($N_{cyl}$). Then we show the necessity of the sketch loss by deactivating it to train the network. 
The quantified results are presented in Table~\ref{table:ablation}. Settings (a), (b), and (c) show that reducing the number of SK-heads or increasing the number of cylinder outputs will damage the model prediction accuracy. Settings (b), (d), and (e) show that increasing the number of MLP layers in SK-head or enabling $\mathcal{L}_{sketch}$ will improve the prediction accuracy.


\begin{table}[]
\caption{Ablation study on sketch representation. }
\label{table:ablation_sk} 
\begin{tabularx}{\columnwidth}{c|Y|Y|Y|Y}
\hline
 Representation & CD↓ & ECD↓ & NC↑ & \#P↓ \\ \hline\hline
Box primitives        & 0.523          & 0.982          & 0.825          & 5.38         \\ \hline
BSP                   & 0.612          & 0.838          & 0.852          & 5.84         \\ \hline
SK-head (Ours) & \textbf{0.330} & \textbf{0.724} & \textbf{0.863} & \textbf{4.30} \\ \hline
\end{tabularx}
\end{table}

\begin{figure}[!t]
    \centerline{
    \includegraphics[width=1.0\linewidth]{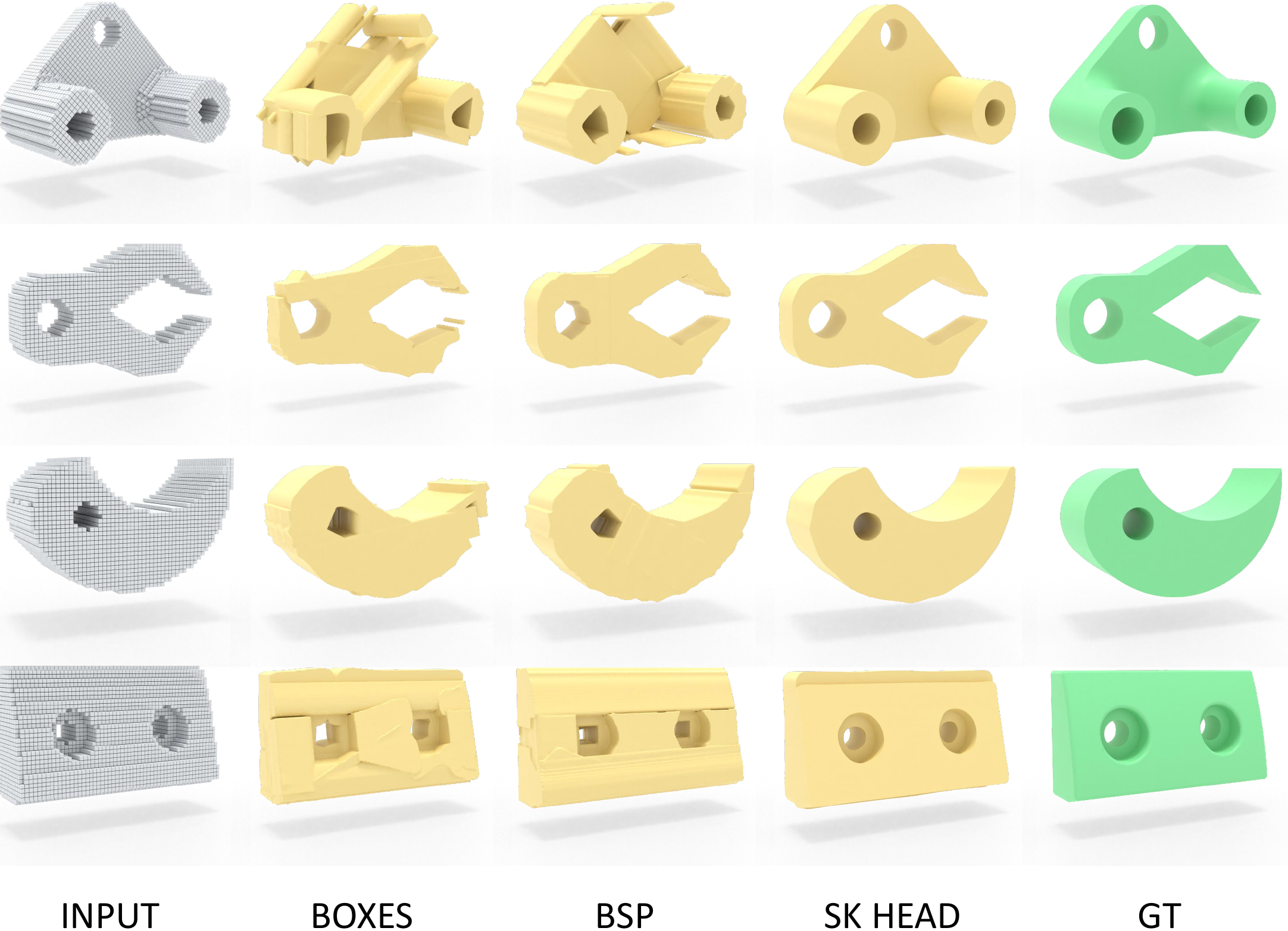}
    }
    \caption{Visual comparison results for ablation study on sketch representation.
    }
    \label{fig:ablation_vis}
\end{figure}

\noindent\textbf{Effect of implicit sketch representation.}
To assess the efficiency of neural implicit sketch representation, we adopt two other classical shape representations, namely binary
space partitioning (BSP~\cite{chen2020bsp}) and box-like primitives, to compare with our SK-head in SECAD-Net. For BSP, we set the number of output convex shapes to 8, each containing 12 partitions. The assembly method is consistent with ~\cite{chen2020bsp} to represent 2D sketches. For box-like primitives, 24 rectangles are predicted. We divide them into two subsets in half, take the union operation separately, and subtract the other from one of the union results. The numerical and visual comparison results are shown in Table~\ref{table:ablation_sk} and Fig.~\ref{fig:ablation_vis}, respectively. It can be seen that our implicit field can represent the smoothest shape while obtaining the best reconstruction results.


\begin{figure}[!t]
    \centerline{
    \includegraphics[width=1.0\linewidth]{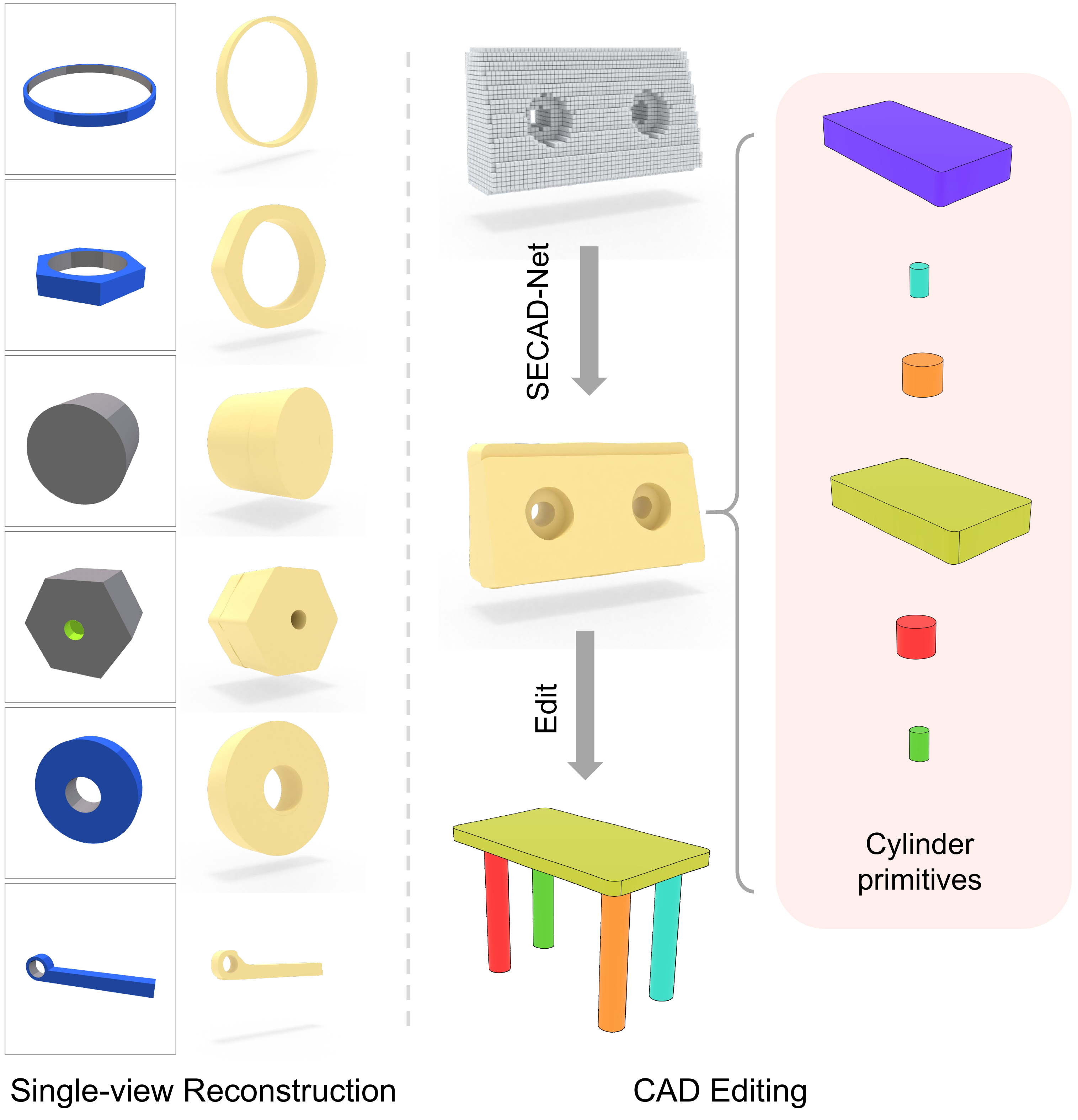}
    }
    \caption{SECAD-Net can aid in more applications. Left: the results of single-view reconstruction. Right: a subsequent CAD editing by changing the predicted cylinder primitives.
    }
    \label{fig:other_app}
\end{figure}

\subsection{Other Applications}

By replacing the voxel encoder, SECAD-Net is flexible to reconstruct CAD models from other input shape representations, \eg, images and point clouds. Fig.~\ref{fig:other_app} shows the results of SECAD-Net in solving single-view reconstruction (SVR) task. Following the training strategy of previous work~\cite{chen2019learning,chen2020bsp}, we first use voxel data to complete the training of the 3D auto-encoding task, and then train an image encoder with the feature encoding of each shape as the target. The voxels and input images used for the SVR task are obtained directly from Fusion 360. Replacing more input representations, while feasible and meaningful, is not the focus of this paper and we leave it to future research.

Finally, the parameters of both 2D sketches and 3D cylinders are available, thus the CAD results output from SECAD-Net can be directly loaded into existing CAD software for further editing. As shown in the right side of Fig.~\ref{fig:other_app}, interpretable CAD variations can be produced via specific editing operations, such as sketch-level curve editing, primitive-level displacement, rotation, scaling, and Boolean operations between primitives. 
\section{Conclusion and Future Work}
\label{sec:conclusion}

We have presented a novel neural network that successively learns shape sketch and extrusion without any expensive annotations of shape segmentation and labels as the supervision.
Our approach is able to learn smooth sketches, followed by the differentiable extrusion to reconstruct CAD models that are close to the ground truth. 
We evaluate SECAD-Net using diverse CAD datasets and demonstrate the advantages of our approach by ablation studies and comparing it to the state-of-the-art methods. 
We further demonstrate our method’s applicability in single-image CAD reconstruction. 
Additionally, the CAD shapes generated by our approach can be directly fed into off-the-shelf CAD software for sketch-level or cylinder primitive-level editing. 



In future work, we plan to extend our approach to learn more CAD-related operations such as \emph{revolve, bevel, and sweep}. 
Besides, we find that current deep learning models perform poorly on datasets with large differences in shape geometry and structure. 
Therefore, another promising direction is to explore how to improve the generalization of neural networks and enhance the realism of the generated shapes by learning structural and topological information.
\\

\noindent\textbf{Acknowledgments.} 
We thank the anonymous reviewer for their valuable suggestions. 
This work is partially funded by the National Natural Science Foundation of China (U22B2034, 62172416, U21A20515, 62172415), and the Youth Innovation Promotion Association of the Chinese Academy of Sciences (2022131).

{\small
\bibliographystyle{ieee_fullname}
\bibliography{egbib}
}

\end{document}


\title{SECAD-Net: Self-Supervised CAD Reconstruction by Learning Sketch-Extrude Operations (Supplementary Materials)}

\author{Pu Li$^{1,2}$ \quad Jianwei Guo$^{1,2}$\thanks{Corresponding author: jianwei.guo@nlpr.ia.ac.cn} \quad Xiaopeng Zhang$^{1,2}$ \quad Dong-Ming Yan$^{1,2}$\\
$^1$MAIS, Institute of Automation, Chinese Academy of Sciences\\
$^2$School of Artificial Intelligence, University of Chinese Academy of Sciences
}

\maketitle

\section{Details of Evaluation Metrics}
In the main paper, we used symmetric Chamfer Distance ($\textit{CD}$), Normal Consistency ($\textit{NC}$), Edge Chamfer Distance ($\textit{ECD}$) as evaluation metrics.  
Specifically, \textit{CD} measures the squared distances between nearest neighbors of two point clouds:
\begin{equation}
CD= \mathbb{E}_{\mathbf{x}  \in \mathbf{P} } \min _{\mathbf{y}  \in \mathbf{G} }\|\mathbf{x} -\mathbf{y} \|_{2}^{2}+ \mathbb{E}_{\mathbf{y}  \in \mathbf{G} } \min _{\mathbf{x}  \in \mathbf{P} }\|\mathbf{x} -\mathbf{y} \|_{2}^{2},
\end{equation}
where $\mathbf{P}$ and $\mathbf{G}$ are the point clouds uniformly sampled from the prediction mesh and ground truth mesh.
\textit{ECD} is calculated in the same way as \textit{CD}, except that the targets are points that lie on the sharp edges of the shape, forming subsets $\mathbf{P}^\mathsf{E}$ and $\mathbf{G}^\mathsf{E}$ of uniformly sampled points:
\begin{equation}
ECD=\mathbb{E}_{\mathbf{x}  \in \mathbf{P}^\mathsf{E}   } \min _{\mathbf{y}  \in \mathbf{G}^\mathsf{E}}\|\mathbf{x} -\mathbf{y} \|_{2}^{2}+\mathbb{E}_{\mathbf{y}  \in \mathbf{G}^\mathsf{E}} \min _{\mathbf{x}  \in \mathbf{P}^\mathsf{E}}\|\mathbf{x} -\mathbf{y} \|_{2}^{2}.
\end{equation}
The close-to-edge points for calculating \textit{ECD} are obtained from 12,000 surface-sampled points. We adopt the settings of distance and sharpness threshold from ~\cite{chen2020bsp}.
Also utilizing the nearest neighbors, \textit{NC} calculates the dot product of their normal vectors:
\begin{equation}
NC=
\left (
\mathbb{E}_{\mathbf{x}  \in \mathbf{P}}\left [\mathbf{n}_{\mathbf{x}}\cdot\mathbf{n}_{\mathbf{y}_\mathbf{x}}\right ]
+\mathbb{E}_{\mathbf{y}  \in \mathbf{G}}\left [\mathbf{n}_{\mathbf{y}}\cdot\mathbf{n}_{\mathbf{x}_\mathbf{y}}\right ]  
\right ) /2
,
\end{equation}
where $\mathbf{n}$ represents the normal vector, $\mathbf{x}_\mathbf{y}$ stands for the neighbor points of point cloud $\mathbf{x}$ in point cloud $\mathbf{y}$, and vice versa. We use 8,192 surface-sampled points to compute \textit{CD} and \textit{NC}. For easy viewing, we scale \textit{CD} by 1,000 and \textit{ECD} by 100.

\section{Discussion of primitives and assemblies of different methods}
For primitive shapes, there is a trade-off between expressiveness and ease of editing. In addition to the methods we compared in the main paper, BSP-Net~\cite{chen2020bsp} and CAPRI-Net~\cite{yu2022capri} are two other methods that use primitive shapes as well as CSG operations for shape reconstruction. 
As shown in Table~\ref{table:primitive}, the primitive shapes (convexes in the original papers) of BSP-Net and CAPRI-Net are intersections of linear partitions or quadratic partitions, which cannot be edited directly in 2D. 
UCSG-Net and CSG-Stump adopt fixed shapes, such as boxes and spheres, as primitives. The outputs of our method, point2Cyl and ExtrudeNet contain sketches, making the cylinder primitives editable in 2D. As shown in Table~\ref{table:primitive}, the property of generating primitives that can be edited in 2D makes the methods in the last four rows within a fair comparison, so we report qualitative and quantitative comparisons of such methods in the main text.

\begin{table}[h]
\caption{Comparison of different shape representations for reconstruction methods based on primitives.}
\label{table:primitive} 
\begin{tabularx}{\columnwidth}{c|Y|c}
\hline
\textbf{Method} & \textbf{Primitives}                   & \textbf{2D editable} \\ \hline
BSP-Net~\cite{chen2020bsp}         & Intersection of linear partitions    & \xmark               \\ \hline
CAPRI-Net~\cite{yu2022capri}       & Intersection of quadratic partitions & \xmark               \\ \hline
UCSG-Net~\cite{kania2020ucsg}        & Box, Sphere                          & \cmark               \\ \hline
CSG-Stump~\cite{ren2021csg}       & Box, Sphere, Cylinder, Cone          & \cmark               \\ \hline
Point2Cyl~\cite{uy2022point2cyl}       & Extrusion cylinder                   & \cmark               \\ \hline
ExtrudeNet~\cite{ren2022extrudenet}       & Extrusion cylinder                   & \cmark               \\ \hline
Ours            & Extrusion cylinder                   & \cmark               \\ \hline 
\end{tabularx}
\end{table}

\section{More comparison results}
We show more reconstruction results of each method on the ABC and Fusion 360 datasets in Fig.~\ref{fig:spp_abc} and Fig.~\ref{fig:spp_fusion}, respectively. The modeling algorithm of the output meshes and the display order of the results are the same as in the main paper. \rev{To highlight the differences between SECAD-Net and ExtrudeNet~\cite{ren2022extrudenet}, we further show the segmentation of the reconstruction results of the two methods in Fig. 1. Each color in the figure represents a primitive before the final union operation. }
\begin{figure}[h]
    \centerline{
    \includegraphics[width=1.0\linewidth]{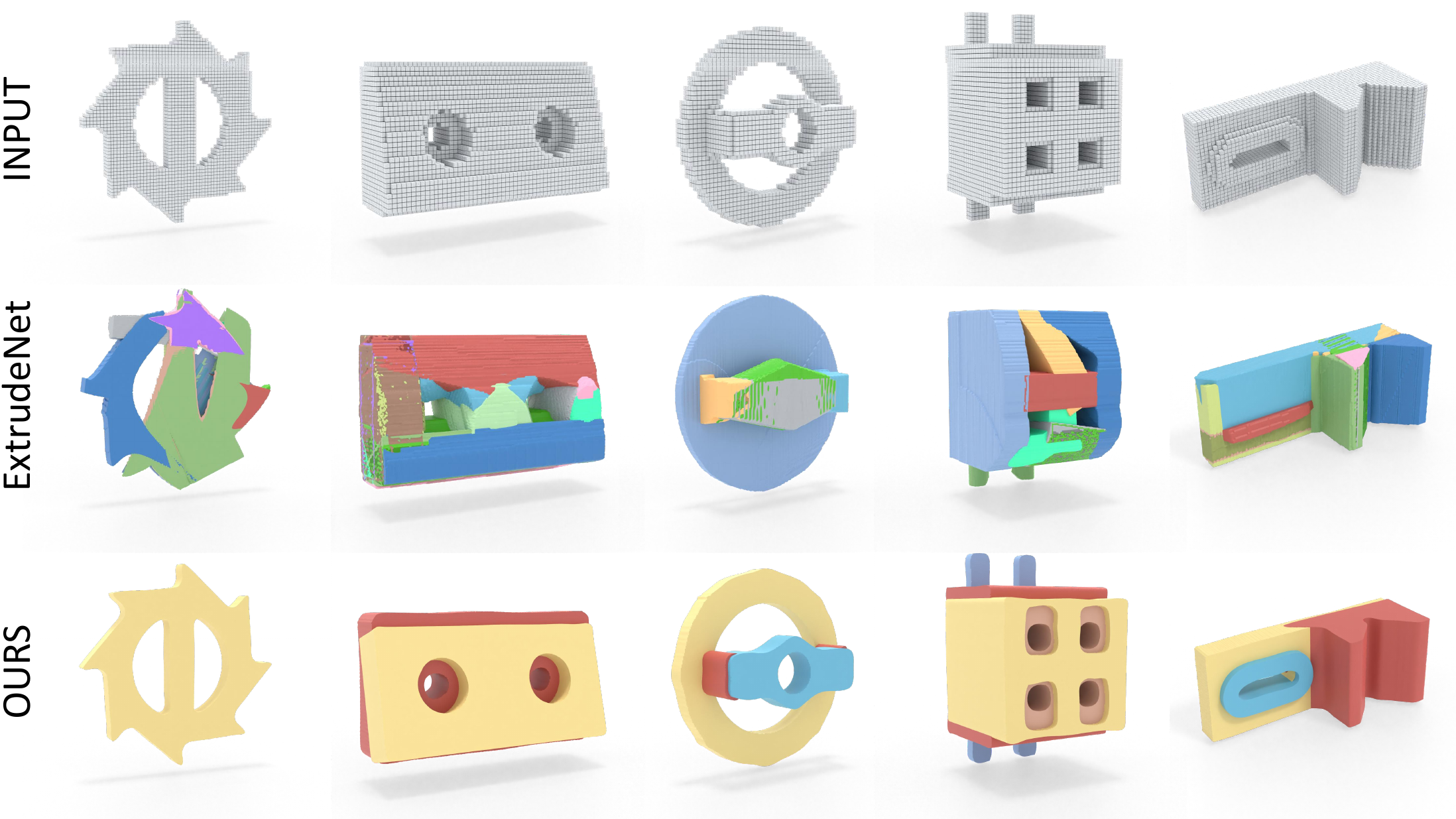}
    }
    \caption{Visual comparison between ExtrudeNet~\cite{ren2022extrudenet}  and ours. The different colors of the parts indicate different primitives. 
    }
    \label{fig:rebuttal_vis}
\end{figure}

\section{Comparison results without fine-tuning}
\rev{We compare the reconstruction performance of each method before fine-tuning on the ABC and Fusion datasets. Using the same evaluation metrics as the main paper, we list the quantitative results in Table~\ref{table:no_finetune}. The results show that our method still achieves the best results in five out of six evaluations.}
\begin{table}[h]
\caption{Quantitative comparison between different methods without fine-tuning on ABC/Fusion 360 datasets. }
\vspace{-1mm}
\label{table:no_finetune} 
\begin{tabularx}{\columnwidth}{E|C|C|D}
\hline
Methods    & CD↓ & ECD↓ & NC↑   \\ \hline \hline
UCSG-Net~\cite{kania2020ucsg}  & 3.14/4.45 & 12.0/17.0 & 70.48/66.63 \\ \hline
CSG-Stumpt~\cite{ren2021csg} & 3.34/5.22 & 3.14/6.88 & 70.70/66.28 \\ \hline
ExtrudeNet~\cite{ren2022extrudenet} & 3.07/5.61 & \textbf{2.99}/6.92 & 71.35/68.15 \\ \hline
Ours & \textbf{2.94}/\textbf{4.20} & 3.00/\textbf{5.53} & \textbf{71.98}/\textbf{68.68} \\ \hline
\end{tabularx}
\end{table}
\section{More ablation comparisons of sketch representations}
In order to demonstrate the advantages of our sketch head network (SK-head) over other classical shape representations, we further show the profiles predicted by different sketch representations in the ablation study in Fig.~\ref{fig:spp_abl_sk}. It can be seen that our SK-head can predict smoother and more complete profiles compared to BSP and box primitives.

\section{More visual results of sketch interpolations}
To further explore the interpretability of the proposed network, we show the 3D shapes and profiles corresponding to the linearly interpolated embedding codes in Fig.~\ref{fig:spp_sk_interp_1} and Fig.~\ref{fig:spp_sk_interp_2}. The new sketches obtained by interpolation can be directly sent to CAD software for editing and re-creation.

\section{CSG assembly using cylinder primitives}
In our final CAD result, the smallest unit is the cylinder primitives, which can be assembled into the final shape using CSG operations. The assembly process forms a CSG tree-like structure, and we show some assembly visualization results in Fig .~\ref{fig:csg}.

\begin{figure*}[!t]
    \centerline{
    \includegraphics[width=\linewidth]{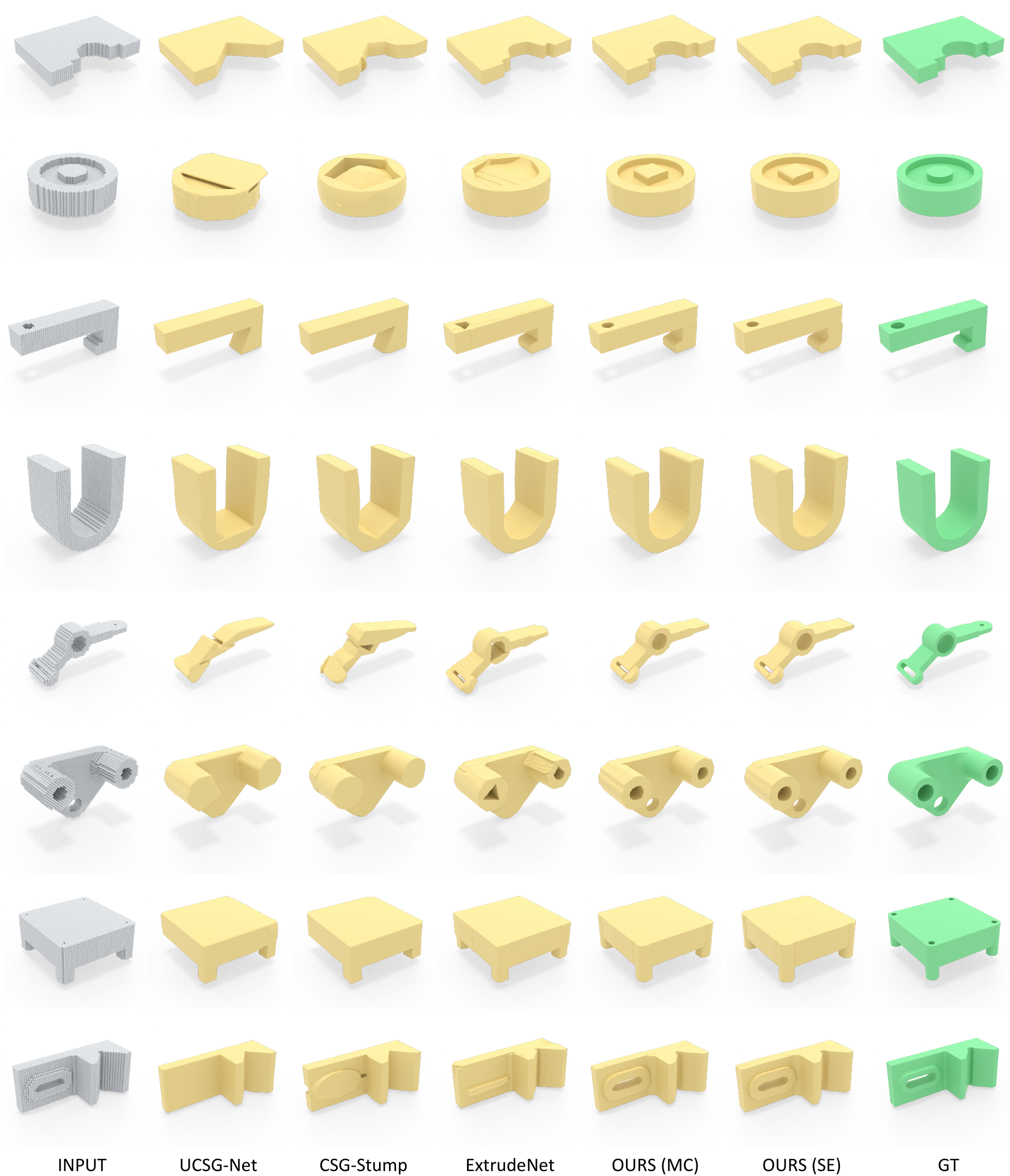}
    }
    \caption{Visual comparison of our approach with UCSG-Net~\cite{kania2020ucsg}, CSG-Stump~\cite{ren2021csg}, and ExtrudeNet~\cite{ren2022extrudenet} on the ABC dataset. Our reconstruction results are generated using both marching cubes (MC) and our proposed sketch-extrude operations (SE). }
    \label{fig:spp_abc}
\end{figure*}

\begin{figure*}[!t]
    \centerline{
    \includegraphics[width=\linewidth]{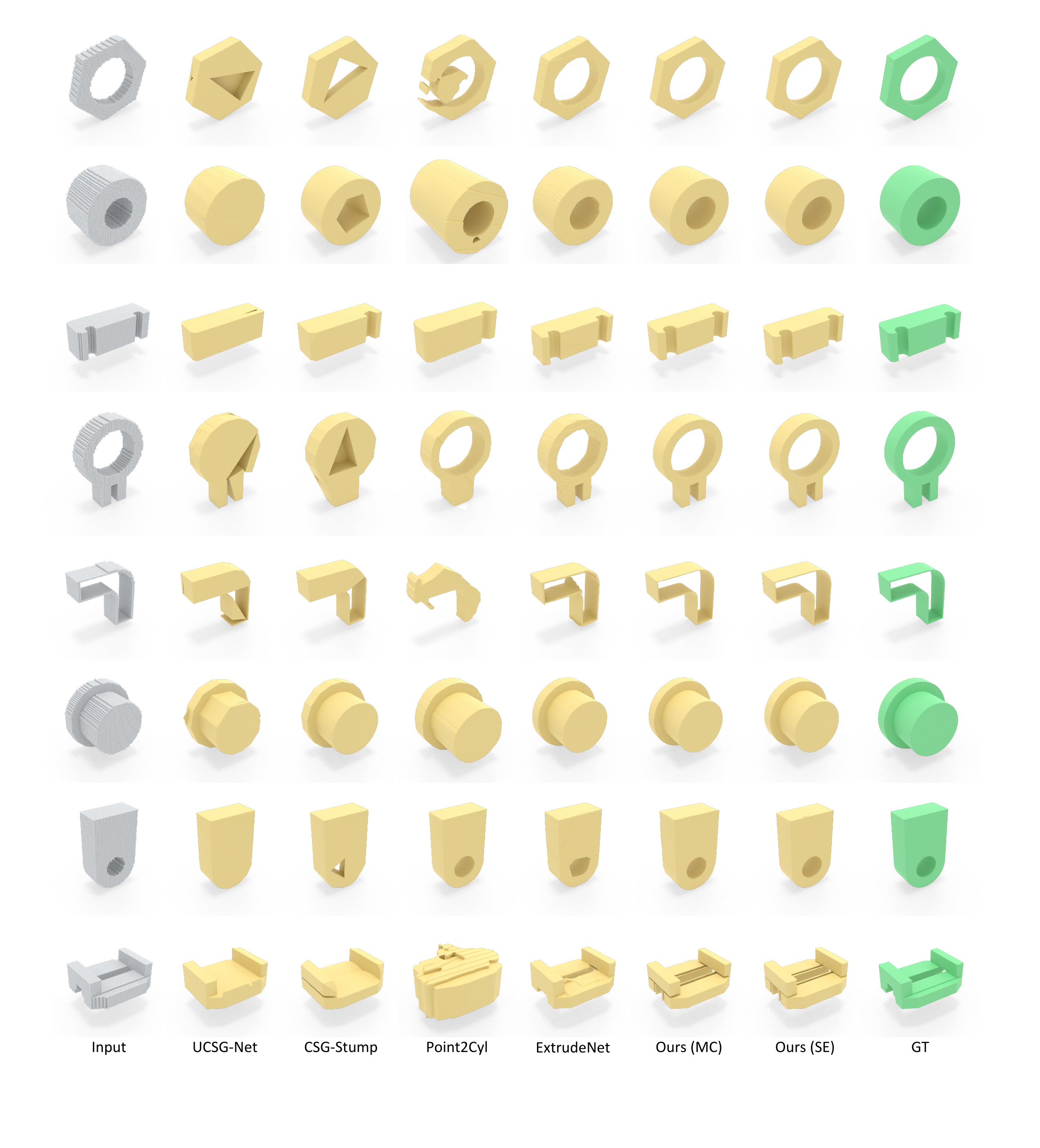}
    }
    \caption{Visual comparison between reconstruction results on Fusion 360 dataset.}
    \label{fig:spp_fusion}
\end{figure*}

\begin{figure*}[!t]
    \centerline{
    \includegraphics[width=1\linewidth]{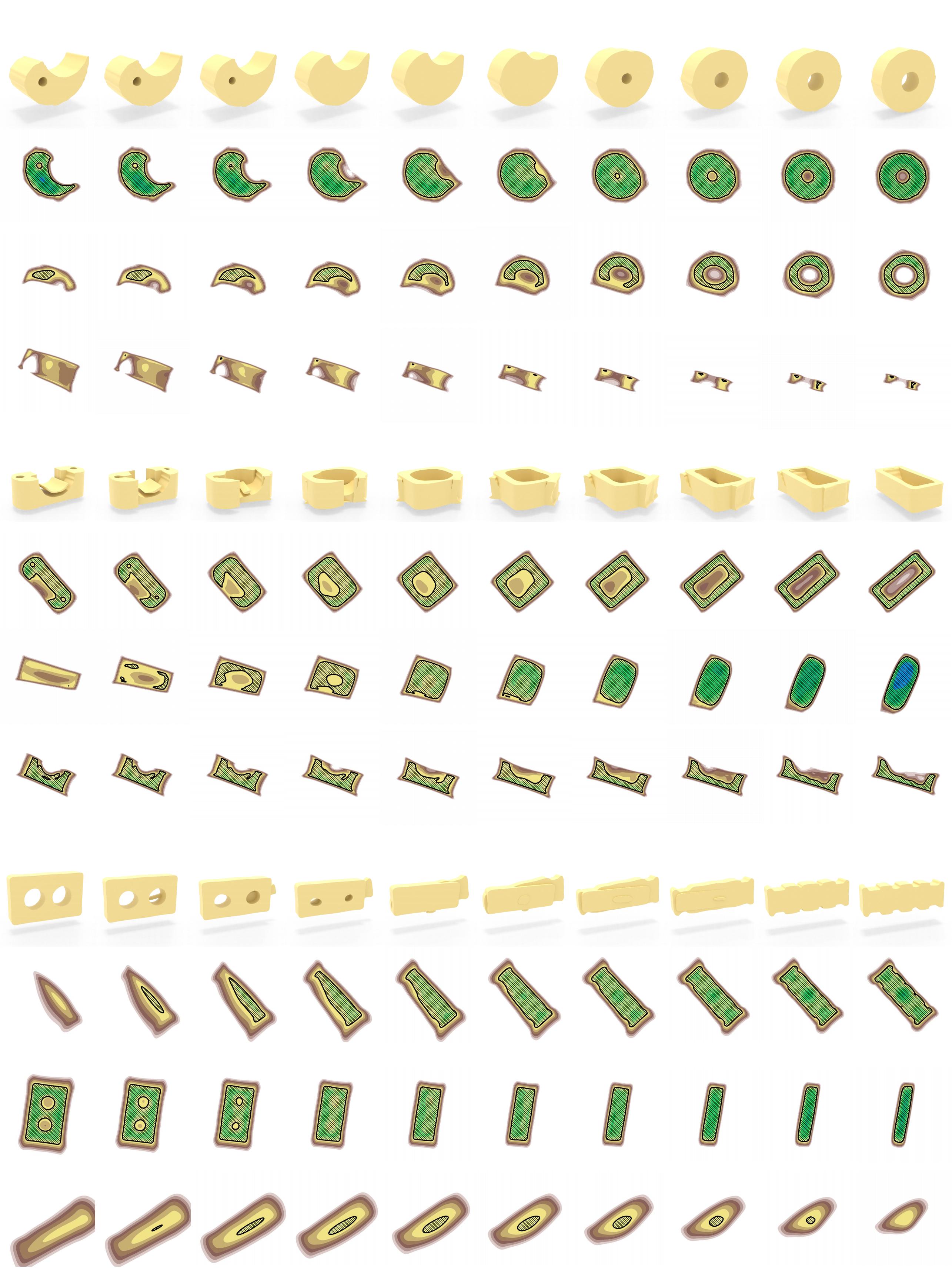}
    }
    \caption{The leftmost and rightmost columns are the prediction results of SECAD-Net on the ABC dataset. The middle results are obtained by interpolating the latent embeddings. For each case, the top row shows the 3D shape, and each remaining row represents the sketches learned by one sketch head network.}
    \label{fig:spp_sk_interp_1}
\end{figure*}

\begin{figure*}[!t]
    \centerline{
    \includegraphics[width=1\linewidth]{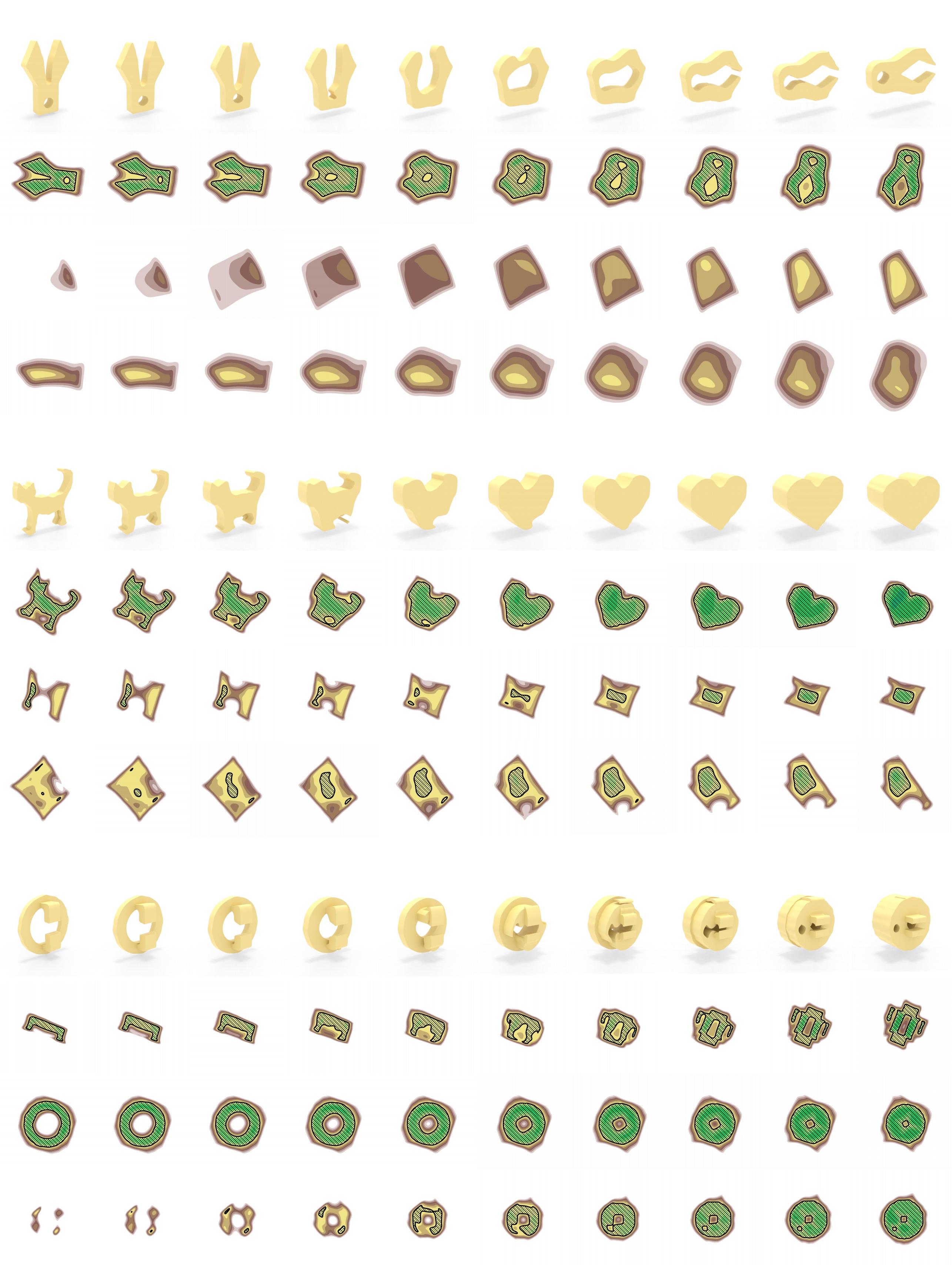}
    }
    \caption{The leftmost and rightmost columns are the prediction results of SECAD-Net on the ABC dataset. The middle results are obtained by interpolating the latent embeddings. For each case, the top row shows the 3D shape, and each remaining row represents the sketches learned by one sketch head network.}
    \label{fig:spp_sk_interp_2}
\end{figure*}

\begin{figure*}[!t]
    \centerline{
    \includegraphics[width=\linewidth]{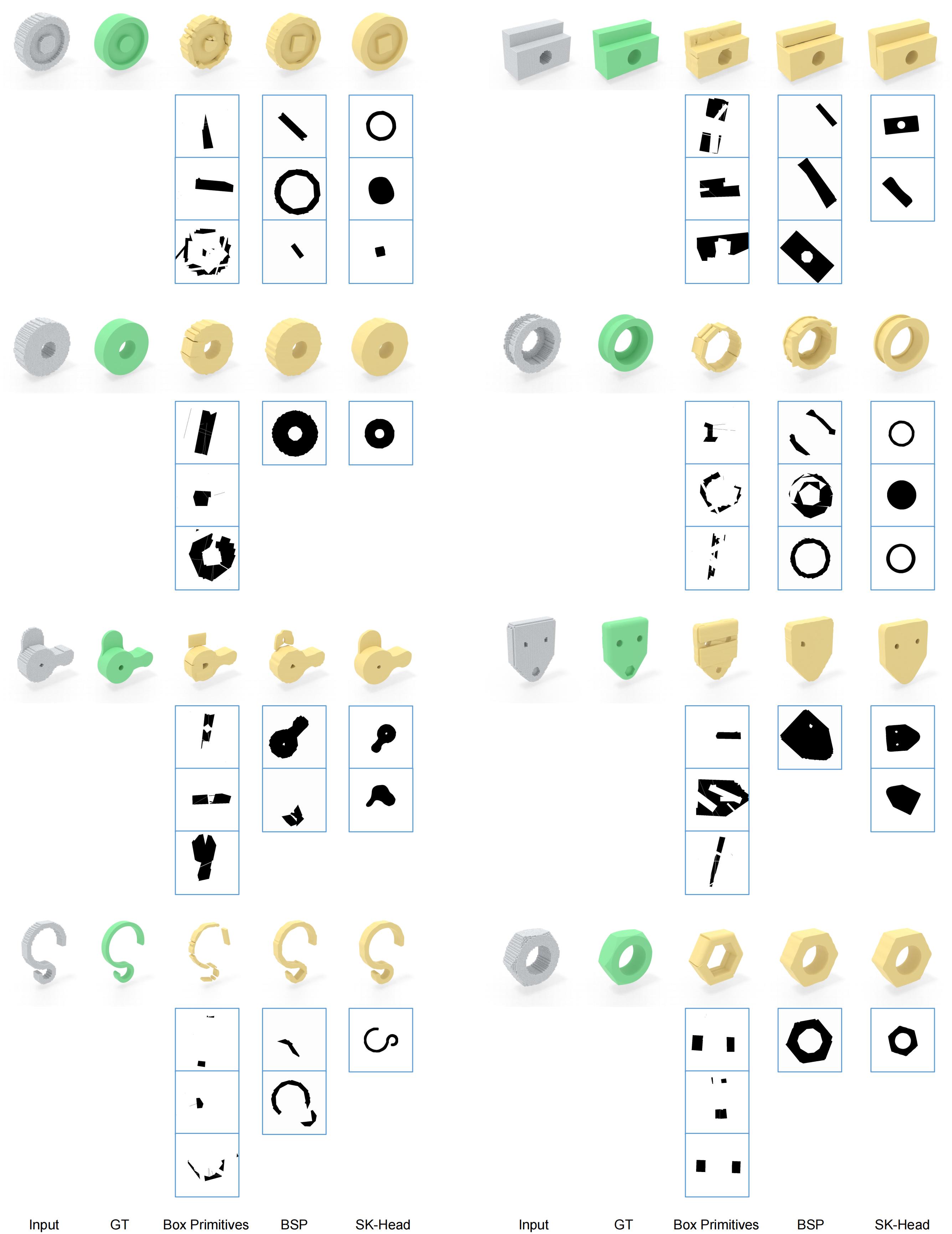}
    }
    \caption{Visual comparison results for ablation study on sketch representation. Each case contains 3D shapes modeled with different sketch representations and the corresponding binary profile images.}
    \label{fig:spp_abl_sk}
\end{figure*}

\begin{figure*}[!t]
    \centerline{
    \includegraphics[width=0.95\linewidth]{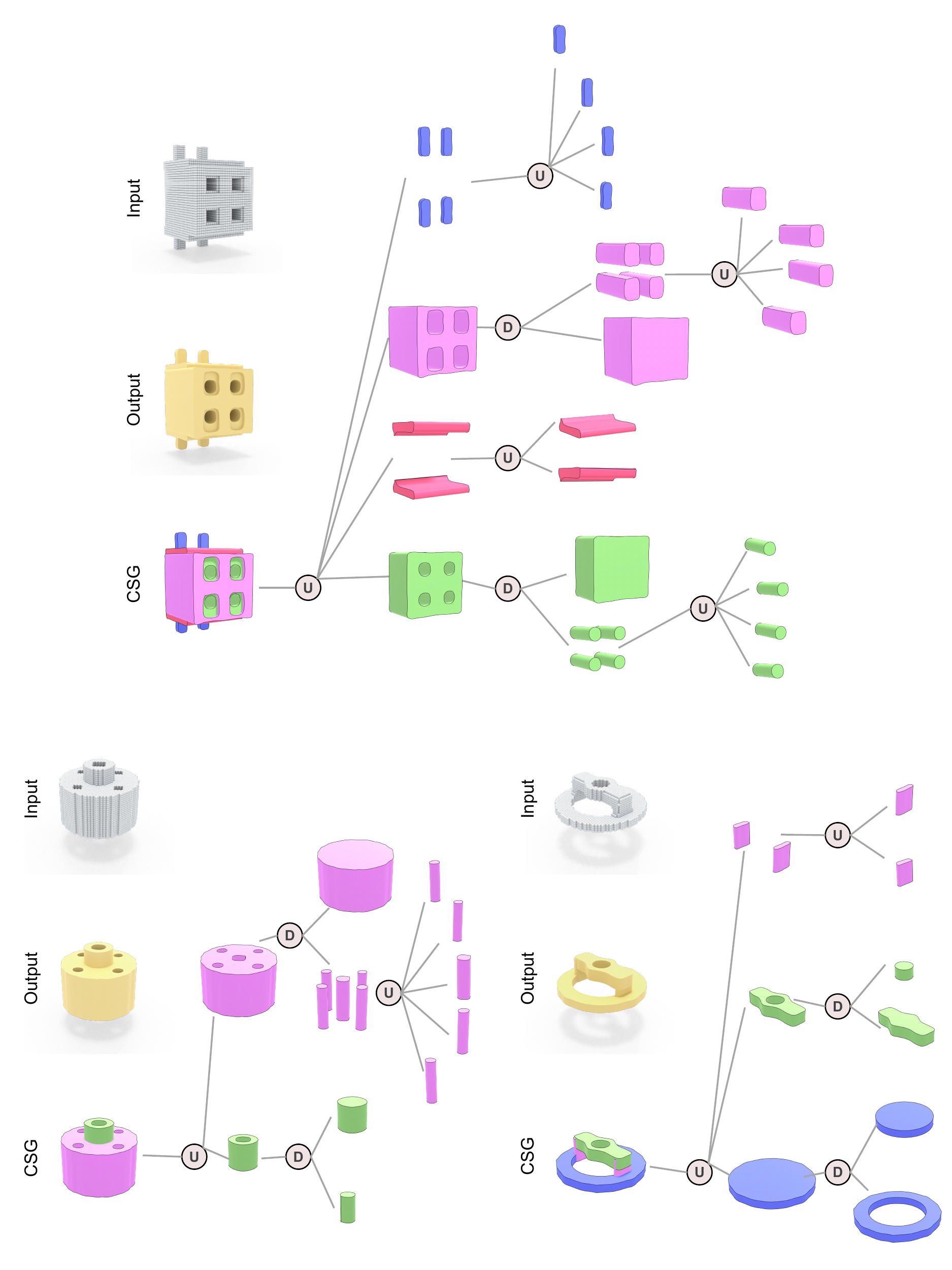}
    }
    \caption{Visualization of assembling cylinder primitives into complete shapes with Boolean operations. Therefore, the final output CAD models are represented as a CSG tree-like structure.}
    \label{fig:csg}
\end{figure*}

{\small
\bibliographystyle{ieee_fullname}
\bibliography{egbib}
}